\documentclass{llncs}

\usepackage{algorithm}
\usepackage{algorithmic}
\usepackage{graphicx}
\usepackage{color}
\usepackage{subfigure}

\renewcommand{\deg}{\theta}
\newcommand{\cp}{curvature point}
\newcommand{\cps}{curvature points}
\newcommand{\fdf}{Fuzzy Directional Features}
\newcommand{\deffdf}{FDF}
\renewcommand{\k}{\beta}

\title{On-line Handwritten Devanagari Character Recognition using \fdf}
\author{Sunil Kumar Kopparapu, Lajish VL\\
\institute{TCS Innovation lab - Mumbai, \\ Yantra Park, Tata Consultancy
Services, Thane (West), Maharastra, India}
\email{SunilKumar.Kopparapu@TCS.Com, Lajish.VL@TCS.Com}}

\begin{document}

\maketitle
\begin{abstract}

This paper describes a new feature set for use in the recognition of 
on-line handwritten Devanagari 
script based on \fdf. Experiments are 
conducted for the automatic recognition of isolated handwritten 
character primitives (sub-character units). Initially we describe the 
proposed feature set, called the \fdf\ (FDF) and 
then show how these features can be effectively utilized for writer 
independent character recognition. Experimental results show that 
FDF set perform well  
for writer independent data set at stroke level recognition. 
The main contribution of this paper is the introduction of a 
novel feature set and establish experimentally its ability in
recognition of handwritten Devanagari script.

\end{abstract}

%
%

\section{Introduction}
\label{sec:intro}

\begin{figure}
\subfigure[Primitive 
that can be used to write the complete
alphabet set in Devanagari.]{
\includegraphics[width=0.40\textwidth]{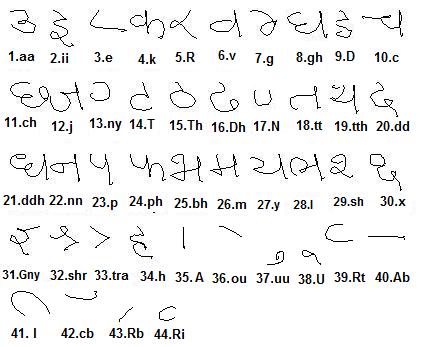}
\label{fig:primitives}
}
\hfill
\subfigure[Variability in writing primitives.] {
\includegraphics[width=0.40\textwidth]{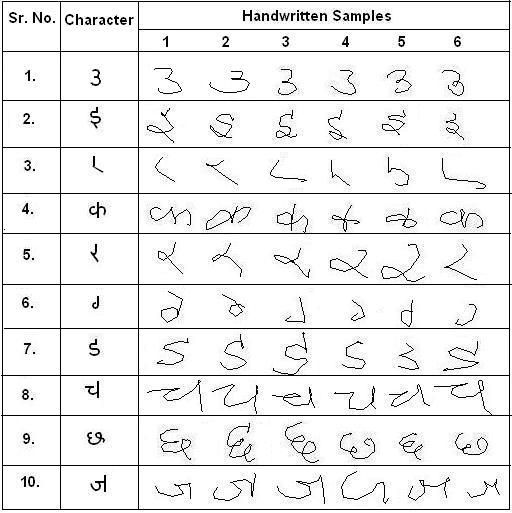}
\label{fig:strokes}
}
\caption{(a) Primitive handwritten strokes, (b) Wide variability is observed}
\end{figure}

Interest in on-line handwritten script recognition has been active for a 
long time. 
In the case of Indian languages, research work is active 
especially for Devanagari \cite{bb98394,bb99164}, Bangala 
\cite{bb98377,bb98378}, Telugu \cite{bb98372} and Tamil 
\cite{bb98368,bb98374} to name a few. 
In English script, the mostly widely researched, barring a few alphabets, 
all the alphabets can be written in a single stroke.
 But most of the Indian languages have characters which are made up 
of two or more strokes which makes it necessary to analyze a set of 
strokes to identify the entire character. 
We identified, through visual inspection of the script, a basis like set of 
$44$ strokes\footnote{ 
Usually the segment of pen motion from 
the pen-down to the pen-up position is a loose definition of a stroke}
called {\em primitives} which are sufficient to represent 
all the characters in Devanagari script.
The set of primitives used to write the complete Devanagari character set 
are shown in Figure \ref{fig:primitives}. In this paper we use these 
primitives 
as the units for recognition. In an 
unconstrained handwriting these primitive strokes exhibit large 
variability in shape, direction and order of writing. 
A sample set of primitives collected from  
different writers 
is shown in Figure \ref{fig:strokes} to capture the variability in the
way primitives are written.
The variations within the 
primitives even for the same writer is evident and it is observed that the
variation among different writers is even larger; making the task of
recognizing these primitives difficult.

The 
main challenge in on-line handwritten character recognition in Indian 
language is the large size of the character set, variation in writing 
style (when the same stroke is written by different writers or the same 
writer at different times) and the similarity between different 
characters in the script.

In this paper, we propose the use of \fdf\ (\deffdf) set for
the recognition of the primitives (which are also strokes). The variations that
exist in the primitives (see Figure \ref{fig:strokes}) test the 
ability of the proposed features to recognize handwritten script. 
The rest of the paper is organized as follows. 
We introduce the \fdf\ 
set in Section \ref{sec:fdf}.
Experimental results 
are outlined in Section \ref{sec:experimental_results},
and conclusions are drawn in Section \ref{sec:conslusions}.

\section{\fdf\ Extraction}
\label{sec:fdf}

Several temporal features have been used for script recognition in 
general\cite{bb100820,bb100778,bb100781,bb100775} and for 
on-line Devanagari script recognition in particular. We 
propose a simple yet effective feature set based on fuzzy 
directional feature set\footnote{Note that 
\cite{1457929} talks of fuzzy feature set for
Devanagari script albeit for offline script}. 
The detailed procedure for obtaining these directional 
features is given below.

\begin{figure}
\subfigure[]{
\includegraphics[width=0.225\textwidth]{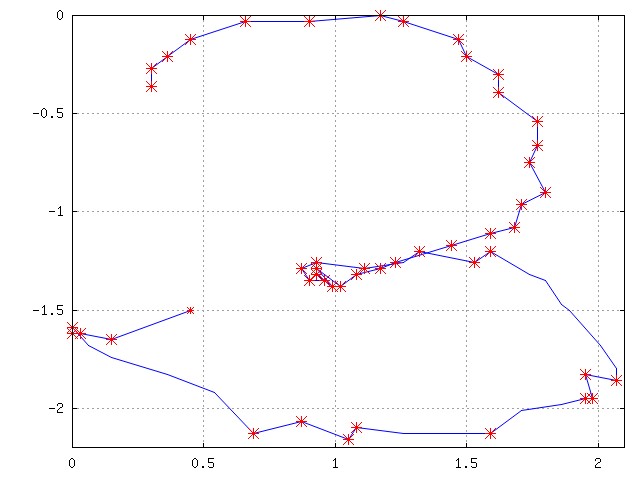}
\label{fig:sampleaa}
}
\subfigure[]{
\includegraphics[width=0.225\textwidth]{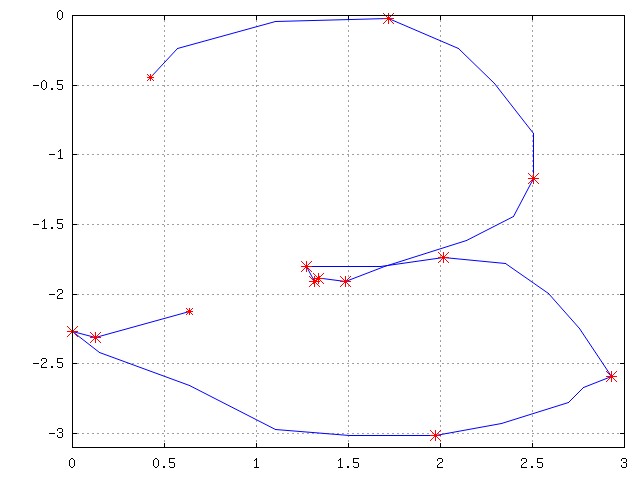}
\label{fig:sampleab}
}
\subfigure[]{
\includegraphics[width=0.225\textwidth]{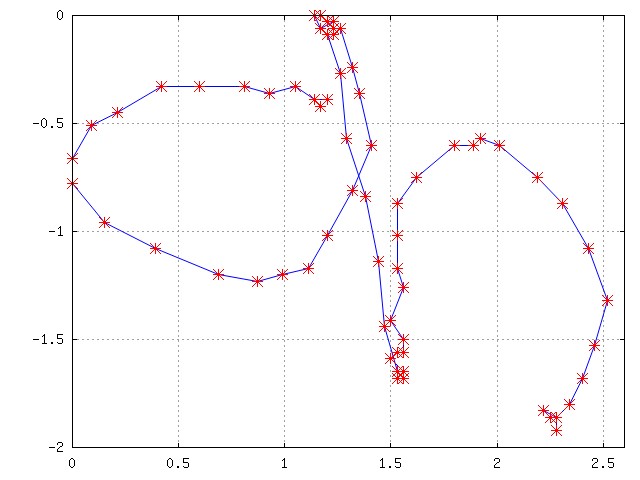}
\label{fig:sampleka}
}
\subfigure[]{
\includegraphics[width=0.225\textwidth]{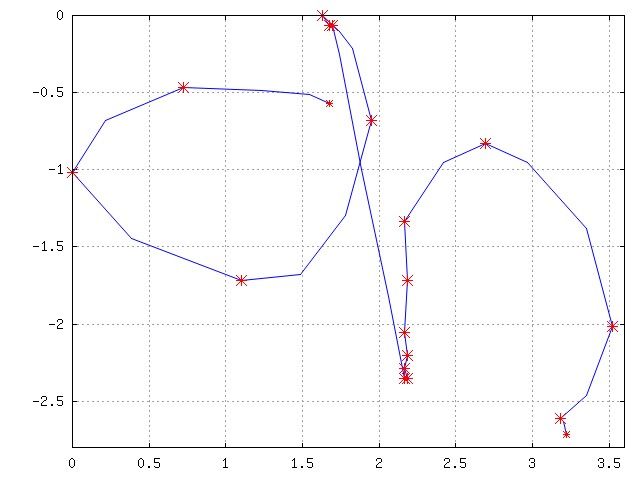}
\label{fig:samplekb}
}
\caption{Sample character (a, c) and its \cps\ (b, d).}
\end{figure}

Let an on-line handwritten character be represented by a variable number of 2D 
points which are in a time sequence. For example an on-line script would 
be represented as
$\left \{(x_{t_1} , y_{t_1} ), (x_{t_2} , y_{t_2} ), \cdots , (x_{t_n} ,
y_{t_n}) \right \} $
where, $t$ denotes the time and $t_1 < t_2 < \cdots < t_n$. 
Equivalently we can represent the on-line character (see Fig 
\ref{fig:sampleaa},\ref{fig:sampleka}) as \[ 
\left \{(x_1 , y_1 ), (x_2 , y_2 ), \cdots, 
(x_n , y_n ) \right \} 
\] 
by dropping the variable $t$. The number of 
points denoted by $n$ vary depending on the size of the character and 
also the speed with which the character is written. 
Most script digitizing devices
(popularly called electronic pen) sample the script uniformly in time,
generally at $100$ Hz. For this 
reason, the number of sampling points is large when the writing speed is 
slow which is especially true at curvatures (see Figure \ref{fig:sampleaa},
\ref{fig:sampleka}); we
exploit these curvature points in extracting \deffdf.

We first identify the curvature 
points (called critical points) from the smoothed ({we use discrete
wavelet transform}) handwriting data. The sequence $(x_i, y_i)_{i=0}^{n}$
represents the handwriting data of a stroke. We treat the sequence $x_i$ and
$y_i$ separately and calculate the critical points for each of these time 
sequence. 
For the $x$ sequence, we calculate the first difference $ x'_i = sgn(x_i -
x_{i+1}) $ 
where $sgn(\k) = +1$ if $\k>0$, $sgn(\k) = -1$ if $\k <0$
and $sgn(\k) = 0$ if $\k =0$. We use $x'$ to compute the critical
point. The point $i$ is a critical point if  $ x'_i - x'_{i+1} \ne 0 $.
Similarly we calculate the critical points for the $y$ sequence. The final list
of critical points is the union of all the points marked as critical points by
both the $x$ and the $y$ sequence (see Figure \ref{fig:sampleab},
\ref{fig:samplekb}).
It must be noted that  the position and number of curvature points computed for
different
samples of the same strokes vary. 
Trimming of \cps\ is carried out on the obtained $k-1$ direction sequence by 
removing all {\em spurious} \cp. A \cp\ is said 
to be spurious if a set of three \cps\ results in the same direction.  For the
sake of discussion lets assume that there are no spurious \cps. 

\begin{figure}
\begin{minipage}[l]{0.35\textwidth}
\subfigure[$\theta$ contributing to two directions ($1$,
$2$) with corresponding membership values (green and 
red dot)]
{\includegraphics[width=0.85\textwidth]{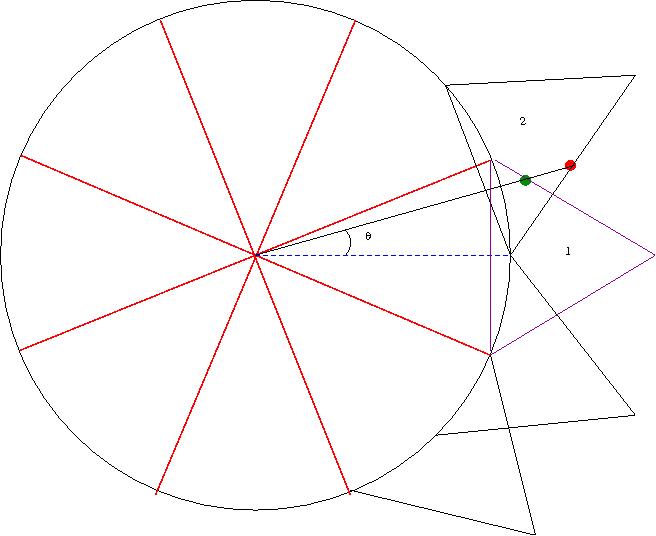}
\label{fig:theta} }
\end{minipage}
\begin{minipage}[l]{0.6\textwidth}
\subfigure[\fdf] {
$
\begin{array}{c|c|c|c|c|c|c|c|c} 
(\theta \downarrow) (d \rightarrow) & 1 & 2 & 3 & 4 & 5&  6&  7&  8 \\ \hline
\theta_{12}&&&m_{12}^1&m_{12}^2&&&& \\
\theta_{23}&m_{23}^1&m_{23}^2&&&&&& \\
\theta_{34}&m_{34}^2&&&&&&&m_{34}^1 \\
\vdots &&&&&&&& \\
\theta_{lm}&&&&&&&m_{lm}^2&m_{lm}^1 \\
\vdots &&&&&&&& \\
\theta_{k-1k}&&&&&m_{k-1k}^2&m_{k-1k}^1&& \\ \hline
\end{array}
$
\label{fig:fdf}
}
\end{minipage}
\caption{\fdf}
\end{figure}

Let $k$ be the number of curvature points
(denoted by $c_1, c_2, \cdots c_k$) extracted from a stroke of length $n$; usually $k << n$.
The $k$ critical points form the basis for extraction of the \deffdf.
We first compute the angle between the two \cps,
say $c_l$ and $c_m$, 
as 
\[ \theta_{lm} = \tan^{-1}\left (\frac{y_l - y_m}{x_l - x_m} \right ) \]
where $(x_l, y_l)$ and $(x_m, y_m)$ are the coordinates corresponding to the
\cp\ $c_l$ and $c_m$ respectively. 
The \deffdf\  set is computed using $\theta_{lm}$. We use
Algorithm \ref{algo:fdf} (assisted by 
triangular membership function, Algorithm \ref{algo:membership}) 
to compute the \deffdf. Note that every $\theta_{lm}$ (represented by $\theta$ in Figure
\ref{fig:theta} is the angle the blue dotted line makes with the $0^o$ axis) 
has two directions
(say $d_{lm}^1 = 1$, $d_{lm}^2 = 2$, note that the line in dotted 
blue in Figure \ref{fig:theta} lies in both 
the triangles represented by direction $1$ and direction $2$) associated with it having 
$m_{lm}^1, m_{lm}^2$ membership values
respectively (represented by the green and the red dot respectively in Figure
\ref{fig:theta}). Also note (a) $m_{lm}^1 + m_{lm}^2 = 1$ and (b) $d_{lm}^1$,
$d_{lm}^2$ are adjacent directions,  for example if $d_{lm}^1 = 5$ then $d_{lm}^2$
could be either $4$ or $6$.

\begin{algorithm}
\begin{algorithmic}
\STATE{deg2fuzzydir(double $\deg$)}
\STATE{i=1; d[i] = -1; m[i] = -1}
\IF{($\deg > -\pi/4$ \& $\deg < \pi/4$)}
	\STATE{d[i] = 1; m[i] = fuzzy\_membership($0$,$\deg$);}
\ENDIF
\IF{($\deg >= 0$ \&  $\deg < 2 \pi/4$) }
	\STATE{d[i] = 2; m[i] = fuzzy\_membership($2 \pi/4$,$\deg$);}
	\STATE{i++;}
\ENDIF
\STATE{\vdots} \COMMENT{{\em{Similarly for d[i] = 3, 4, 5, 6, 7}}}
\IF{($\deg > -2 \pi/4$ \& $\deg < 0$) }
	\STATE{d[i] = 8; m[i] = fuzzy\_membership($0$,$\deg$);}
	\STATE{i++;}
\ENDIF
\STATE{return(d[i], m[i]);}
\end{algorithmic}
\caption{Computing \fdf}
\label{algo:fdf}
\end{algorithm}

\begin{algorithm}
\begin{algorithmic}
\STATE{fuzzy\_membership($\deg_c$, $\deg$)}
\STATE{$m = 1.0 - \frac{(|(\deg_c-\deg)|)}{(\pi/4)}$;}
\STATE{return(m)}
\end{algorithmic}
\caption{Triangular Fuzzy Membership Function}
\label{algo:membership}
\end{algorithm}

It should be noted that the sum
of the membership functions of a particular row (see Figure \ref{fig:fdf}) 
is always $1$.
Given an on-line character, we extract the \deffdf\ shown in
Figure \ref{fig:fdf}.  Then we calculate the mean FDF by averaging across the
columns, so as to form a vector of dimension $8$. The mean is calculated as
follows;  for each direction ($1$ to $8$), collect all the membership values
and divide by the number of occurrences of the membership values in that
direction. For example, for Figure \ref{fig:fdf}, the mean for direction $1$ is
calculated as $f_1 = \frac{(m_{23}^1+m_{34}^1)}{2}$. 
In all our experiments we have
used this mean \deffdf 
\begin{equation}
 {\cal F} = [ f_1, f_2, \cdots f_8 ] 
\label{eq:mfdf}
\end{equation}
to represent a stroke.

\section{Experimental Analysis}
\label{sec:experimental_results}

For experimental analysis, we collected handwritten 
data from $10$ persons, each of whom
wrote all the primitives of Devanagari text using 
Mobile e-Notes
Taker\footnote{http://www.hitech-in.com/mobile\_e-note\_taker.htm}.
This raw stroke data is smoothed
using Discrete Wavelet 
Transform (DWT) decomposition\footnote{We do not dwell on this 
since this is well covered in pattern recognition literature.} to remove noise in terms of small undulation due
to the sensitiveness of the sensors on the electronic pen. 
For each stroke we extracted the fuzzy directional feature set as described
in Section \ref{sec:fdf}.
We used $5$ user data for training and the other $5$ for the purpose 
of testing
the performance of the \deffdf\  set. We initially hand tagged each stroke
in the collected data using the $44$ primitives that we selected (see Figure
\ref{fig:primitives}). 

For training, we calculated (\ref{eq:mfdf}) for  
all strokes corresponding to the same primitive and computed the
average to model the primitive. So a primitive was represented by a vector of size $8$ by taking 
the average over all the occurrences of the primitive in the training set. 
All the experimental results are based
on this data set (from $10$ different writers). 

For testing purpose, we took a stroke ($t$) to be recognized, we first
extracted FDF (using Algorithm \ref{algo:fdf}) and computed the mean \deffdf\
using (\ref{eq:mfdf}). Then we compared it with
the \deffdf\ model of the $44$ reference strokes using the usual Euclidean distance
measure. We computed for the test stroke $t$, its distance from all the
primitives, namely, $|| F_t - F_i ||^2$ for $i = 1, \cdots 44$ and
arranged them in the increasing order of magnitude (best match first). 
The results for this are shown in Table \ref{tab:results} 
for both the train data and the test data for $\alpha =1, 2, 5$. 
Note that the values in Table \ref{tab:results} are computed as followѕ. 
For $N=\alpha$, 
the test stroke $t$ is recognized as the primitive $l$ if 
$l$ occurs atleast at the $\alpha^{th}$ position from the best match 
(this is generally called the N-best in literature).
As expected the recognition
accuracies are poor (very similar to the phoneme recognition by a speech
engine)
for $\alpha=1$ and improves with increasing $\alpha$.
It should be noted that
the accuracies are writer independent and for stroke level recognition.

\begin{table}
\caption{Recognition accuracies for train and test data set.}
\label{tab:results}
\begin{center}
\begin{tabular}{||l|c|c|c||} \hline
Data & $\alpha=1$ & $\alpha=2$ & $\alpha=5$ \\ \hline
Train Data & 63.0\% (139/220)&87.9\% (193/220)& 93.3\%(205/220) \\ \hline
Test Data & 37.0\% (82/220)&54.6\% (120/220) & 78.1\% (172/220)\\ \hline
\end{tabular}
\end{center}
\end{table}
\section{Conclusions}
\label{sec:conslusions}

In this paper we have introduces a new on-line script feature set, 
called the \fdf. 
We have evaluated the performance of the novel feature set by
presented recognition accuracies for writer independent stroke level data set. 
It is
well known, both in speech and script recognition literature that stroke
(phoneme in case of speech) recognition is always poor. As in speech we plan
use (a) Viterbi traceback to enhance alphabet (multiple stroke) recognition 
and/or
(b) cluster strokes using spatio-temporal information 
to form alphabets and then
use the cluster of strokes to recognize them. This we believe
will lead to better accuracies of writer independent script recognition. 

\bibliographystyle{splncs}
\bibliography{script_stat_analysis} 

\begin{thebibliography}{10}
\providecommand{\url}[1]{\texttt{#1}}
\providecommand{\urlprefix}{URL }

\bibitem{bb98372}
Babu, V., Prasanth, L., Sharma, R., Rao, G., Bharath, A.: {HMM}-based online
  handwriting recognition system for {T}elugu symbols. In: ICDAR07. pp. 63--67
  (2007)

\bibitem{bb98374}
Bharath, A., Madhvanath, S.: Hidden markov models for online handwritten
  {T}amil word recognition. In: ICDAR07. pp. 506--510 (2007)

\bibitem{bb98378}
Bhattacharya, U., Gupta, B., Parui, S.: Direction code based features for
  recognition of online handwritten characters of {B}angla. In: ICDAR07. pp.
  58--62 (2007)

\bibitem{bb100775}
Connell, S., Jain, A.: Template-based online character recognition. Pattern
  Recognition  34(1),  1--14 (January 2001)

\bibitem{bb100778}
Garcia~Salicetti, S., Dorizzi, B., Gallinari, P., Wimmer, Z.: Maximum mutual
  information training for an online neural predictive handwritten word
  recognition system. IJDAR  4(1),  56--68 (2001)

\bibitem{bb98394}
Joshi, N., Sita, G., Ramakrishnan, A., Deepu, V., Madhvanath, S.: Machine
  recognition of online handwritten {D}evanagari characters. In: ICDAR05. pp.
  II: 1156--1160 (2005)

\bibitem{bb100820}
Menier, G., Lorette, G., Gentric, P.: A genetic algorithm for on-line cursive
  handwriting recognition. In: ICPR. pp. B:522--525 (1994)

\bibitem{1457929}
Mukherji, P., Rege, P.P.: Fuzzy stroke analysis of devnagari handwritten
  characters. W. Trans. on Comp.  7(5),  351--362 (2008)

\bibitem{bb99164}
Namboodiri, A., Jain, A.: Online handwritten script recognition. PAMI  26(1),
  124--130 (January 2004)

\bibitem{bb98377}
Parui, S., Guin, K., Bhattacharya, U., Chaudhuri, B.: Online handwritten
  {B}angla character recognition using {HMM}. In: ICPR08. pp. 1--4 (2008)

\bibitem{bb100781}
Schenk, J., Rigoll, G.: Neural net vector quantizers for discrete hmm-based
  on-line handwritten whiteboard-note recognition. In: ICPR. pp. 1--4 (2008)

\bibitem{bb98368}
Sundaram, S., Ramakrishnan, A.: A novel hierarchical classification scheme for
  online {T}amil character recognition. In: ICDAR07. pp. 1218--1222 (2007)

\end{thebibliography}

\end{document}